%% file: 0-main.tex
\documentclass[11pt,a4paper]{article}
\usepackage[hyperref]{acl2020}
\usepackage{times}
\usepackage{latexsym}

\usepackage{microtype}

\usepackage{booktabs}
\usepackage{xspace}
\usepackage{arydshln}
\usepackage{graphics}
\usepackage{graphicx}
\usepackage{amsmath}
\usepackage{enumitem}
\usepackage{caption}
\usepackage{subcaption}
\usepackage{multirow}
\usepackage{centernot}
\usepackage[normalem]{ulem}

\newcommand{\calP}{\mathcal{P}}
\newcommand{\calL}{\mathcal{L}}
\newcommand{\cls}{\texttt{[CLS]}\xspace}
\newcommand{\V}[1][\mathbf]{#1}

\newcommand{\sys}{\textsc{Specter}\xspace}
\newcommand{\dataset}{\textsc{SciDocs}\xspace}
\newcommand{\bert}{BERT\xspace}
\newcommand{\pos}{\calP^{+}}
\newcommand{\negative}{\calP^{-}}
\newcommand{\query}{\calP^{Q}}
\newcommand{\map}{\textsc{map}\xspace}
\newcommand{\ndcg}{n\textsc{dcg}\xspace}

\DeclareMathOperator{\dis}{d}
\aclfinalcopy %
\def\aclpaperid{429} %

\title{SPECTER: Document-level Representation Learning using \\ Citation-informed Transformers}

\author{Arman Cohan$^\dag$\thanks{\hspace{0.6em}Equal contribution} \hspace{0.8em} Sergey Feldman$^\dag$\textcolor{darkblue}{\footnotemark[1]} \hspace{0.8em} Iz Beltagy$^\dag$ \hspace{0.8em} Doug Downey$^\dag$ \hspace{0.8em}  Daniel S. Weld$^{\dag,\ddag}$\vspace{8pt} \\ 
  $^\dag$Allen Institute for Artificial Intelligence \vspace{2pt} \\
  $^\ddag$Paul G. Allen School of Computer Science \& Engineering, University of Washington \vspace{2pt}\\
  \texttt{\{armanc,sergey,beltagy,dougd,danw\}@allenai.org}}

\date{}

\makeatletter
\newcommand\cdotfill{%
  \leavevmode\cleaders\hb@xt@.2em{\hss\tiny{$\cdot$}\hss}\hfill\kern\z@
}
\makeatother

\begin{document}
\maketitle
\begin{abstract}
Representation learning is a critical ingredient for natural language processing systems.  Recent Transformer language models like BERT learn powerful textual representations, but these models are targeted towards token- and sentence-level training objectives and do not leverage information on inter-document relatedness, which limits their document-level representation power.
For applications on scientific documents, such as classification and recommendation, the embeddings power strong performance on end tasks. We propose \sys, a new method to generate document-level embedding of scientific documents based on pretraining a Transformer language model on a powerful signal of document-level relatedness: the citation graph. Unlike existing pretrained language models, \sys can be easily applied to downstream applications without task-specific fine-tuning.
Additionally, to encourage further research on document-level models, we introduce \dataset, a new evaluation benchmark consisting of seven document-level tasks ranging from citation prediction, to document classification and recommendation. We show that \sys outperforms a variety of competitive baselines on the benchmark.\footnote{ \url{https://github.com/allenai/specter}} 
\end{abstract}

\input{1-intro.tex}

\input{2-model.tex}

\input{3-exp.tex}

\input{4-results.tex}

\input{5-analysis.tex}

\input{6-related.tex}

\section{Conclusions and Future Work}

We present \sys, a model for learning representations of scientific papers, based on a Transformer language model that is pretrained on citations. We achieve substantial improvements over the strongest of a wide variety of baselines, demonstrating the effectiveness of our model. We additionally introduce \dataset, a new evaluation suite consisting of seven document-level tasks and release the corresponding datasets to foster further research in this area. 

The landscape of Transformer language models is rapidly changing and newer and larger models are frequently introduced. It would be interesting to initialize our model weights from more recent Transformer models to investigate if additional gains are possible. Another item of future work is to develop better multitask approaches to leverage multiple signals of relatedness information during training. 
We used citations to build triplets for our loss function, however there are other metrics that have good support from the bibliometrics literature \cite{Klavans2006IdentifyingAB} that warrant exploring as a way to create relatedness graphs. Including other information such as outgoing citations as additional input to the model would be yet another area to explore in future.

\section*{Acknowledgements}

We thank Kyle Lo, Daniel King and Oren Etzioni for helpful research discussions, Russel Reas for setting up the public API, Field Cady for help in initial data collection and the anonymous reviewers (especially Reviewer 1) for comments and suggestions. This work was supported in part by NSF Convergence Accelerator award 1936940, ONR grant N00014-18-1-2193, and the University of Washington WRF/Cable Professorship.

\bibliography{acl2020}
\bibliographystyle{acl_natbib}

\appendix
\input{9-appendix.tex}

\end{document}

%% file: 1-intro.tex
\section{Introduction}
As the pace of scientific publication continues to increase, Natural Language Processing (NLP) tools that help users to search, discover and understand the scientific literature have become critical. In recent years, substantial improvements in NLP tools have been brought about by pretrained neural language models (LMs) \cite{radford2018improving,Devlin2018BERTPO,Yang2019XLNetGA}.   While such models are widely used for representing individual words or sentences, extensions to whole-document embeddings are relatively underexplored.  Likewise, methods that do use inter-document signals to produce whole-document embeddings \cite{Tu2017CANECN,Chen2019ImprovingTN} have yet to incorporate state-of-the-art pretrained LMs.  Here, we study how to leverage the power of pretrained language models to learn embeddings for scientific documents.

A paper's title and abstract provide rich semantic content about the paper, but, as we show in this work, simply passing these textual fields to an ``off-the-shelf'' pretrained language model---even a state-of-the-art model tailored to scientific text like the recent SciBERT \cite{Beltagy2019SciBERT}---does {\em not} result in accurate paper representations.  The language modeling objectives used to pretrain the model do not lead it to output representations that are helpful for document-level tasks such as topic classification or recommendation.

In this paper, we introduce a new method for learning general-purpose vector representations of scientific documents. Our system, \sys,\footnote{\sys: Scientific Paper Embeddings using Citation-informed TransformERs} incorporates inter-document context into the Transformer \cite{Vaswani2017AttentionIA} language models (e.g., SciBERT \cite{Beltagy2019SciBERT}) to learn document representations that are effective across a wide-variety of downstream tasks, without the need for any task-specific fine-tuning of the pretrained language model. 
We specifically use citations as a naturally occurring, inter-document incidental supervision signal indicating which documents are most related and formulate the signal into a triplet-loss pretraining objective. 
Unlike many prior works, at inference time, our model does not require any citation information. This is critical for embedding new papers that have not yet been cited. In experiments, we show that \sys 's representations substantially outperform the state-of-the-art on a variety of document-level tasks, including topic classification, citation prediction, and recommendation.

As an additional contribution of this work, we introduce and release \dataset\footnote{\url{https://github.com/allenai/scidocs}}  , a novel collection of data sets and an evaluation suite for document-level embeddings in the scientific domain.  \dataset covers seven tasks, and includes tens of thousands of examples of anonymized user signals of document relatedness.  We also release our training set (hundreds of thousands of paper titles, abstracts and citations), along with our trained embedding model and its associated code base.

%% file: 2-model.tex
\section{Model}

\subsection{Overview}
Our goal is to learn task-independent representations of academic papers. Inspired by the recent success of pretrained Transformer language models across various NLP tasks, we use the Transformer model architecture as basis of encoding the input paper. 
Existing LMs such as BERT, however, are primarily based on masked language modeling objective, only considering intra-document context and do not use any inter-document information. This limits their ability to learn optimal document representations.
To learn high-quality document-level representations we propose using citations as an inter-document relatedness signal and formulate it as a triplet loss learning objective. We then pretrain the model on a large corpus of citations using this objective, encouraging it to output representations that are more similar for papers that share a citation link than for those that do not. We call our model \sys, which learns Scientific Paper Embeddings using Citation-informed TransformERs.
With respect to the terminology used by \citet{Devlin2018BERTPO}, unlike most existing LMs that are ``fine-tuning based'', our approach results in embeddings that can be applied to downstream tasks in a ``feature-based'' fashion, meaning the learned paper embeddings can be easily used as features, with no need for further task-specific fine-tuning. In the following, as background information, we briefly describe how pretrained LMs can be applied for document representation and then discuss the details of \sys.

\subsection{Background: Pretrained Transformers}
Recently, pretrained Transformer networks have demonstrated success on various NLP tasks \cite{radford2018improving,Devlin2018BERTPO,Yang2019XLNetGA,Liu2019RoBERTaAR}; we use these models as the foundation for \sys. Specifically, we use SciBERT \cite{Beltagy2019SciBERT} which is an adaptation of the original \bert \cite{Devlin2018BERTPO} architecture to the scientific domain. 
The \bert model architecture \cite{Devlin2018BERTPO} uses multiple layers of Transformers \cite{Vaswani2017AttentionIA} to encode the tokens in a given input sequence. Each layer consists of a self-attention sublayer followed by a feedforward sublayer.
The final hidden state associated with the special {\tt{[CLS]}} token is usually called the ``pooled output'', and is commonly used as an aggregate representation of the sequence.

\begin{figure}
    \centering
    \includegraphics[width=0.8\linewidth]{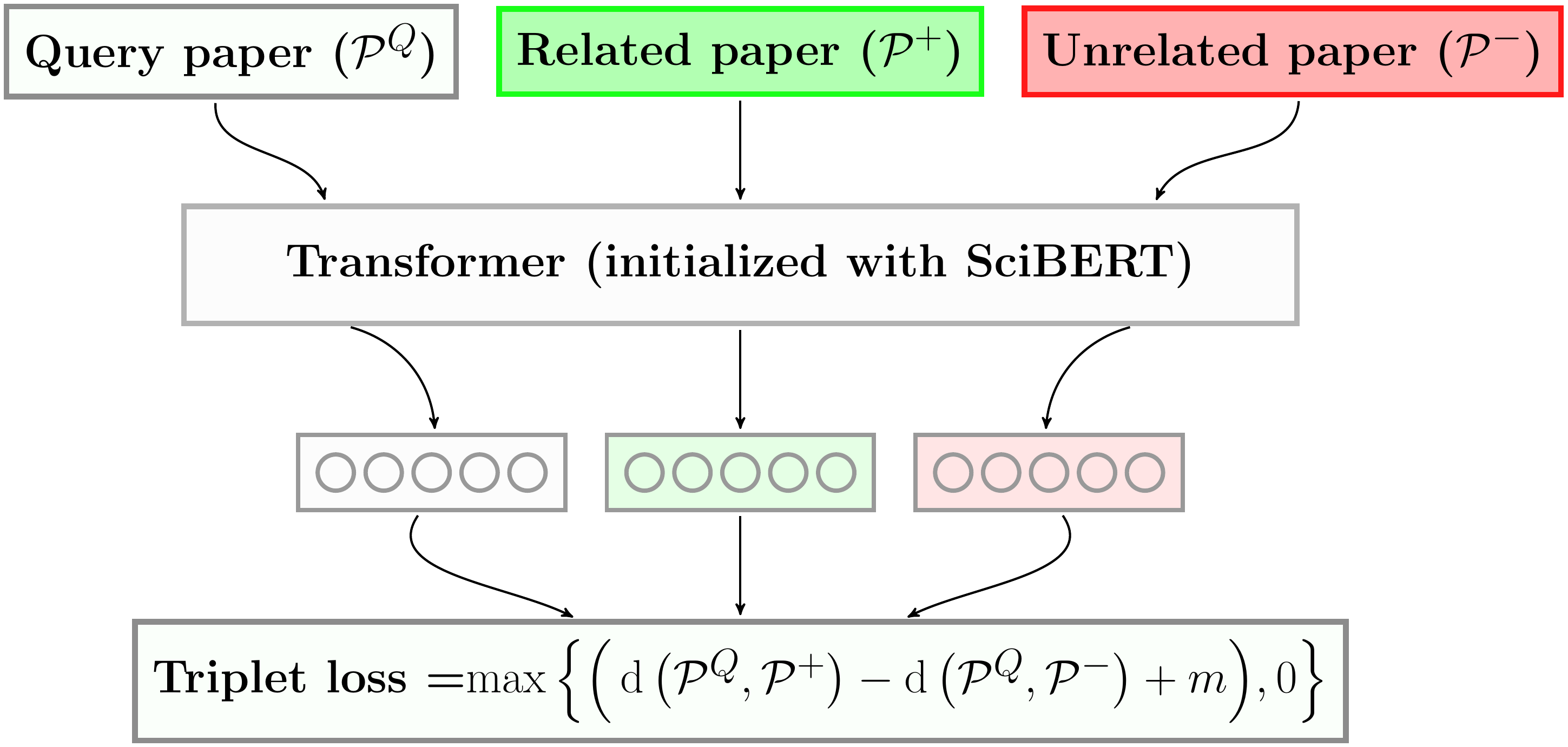}
    \caption{Overview of \sys.}
    \label{fig:overview}
\end{figure}

\paragraph{Document Representation}
\setlength{\abovedisplayskip}{2pt}
\setlength{\belowdisplayskip}{5pt}
Our goal is to represent a given paper $\calP$ as a dense vector $\V{v}$ that best represents the paper and can be used in downstream tasks.
\sys\ builds embeddings from the title and abstract of a paper.
Intuitively, we would expect these fields to be sufficient to produce accurate embeddings,
since they are written to provide a succinct and comprehensive summary of the paper.\footnote{We also experimented with additional fields such as venues and authors but did not find any empirical advantage in using those (see \S\ref{sec:analysis}). See \S\ref{sec:related} for a discussion of using the full text of the paper as input.}  As such, we encode the concatenated title and abstract using a Transformer LM (e.g., SciBERT) and take the final representation of the \cls token as the output representation of the paper:\footnote{It is also possible to encode title and abstracts individually and then concatenate or combine them to get the final embedding. However, in our experiments this resulted in sub-optimal performance.}
\begin{equation}
\V{v} = \mathrm{\texttt{Transformer}}(\texttt{input})_\texttt{[CLS]},
\label{eq:forward-func}
\end{equation}

\noindent where $\mathrm{\texttt{Transformer}}$ is the Transformer's forward function, and \texttt{input} is the concatenation of the \texttt{[CLS]} token and  WordPieces \cite{wu2016google} of the title and abstract of a paper, separated by the \texttt{[SEP]} token. We use SciBERT as our model initialization as it is optimized for scientific text, though our formulation is general and any Transformer language model instead of SciBERT. Using the above method with an ``off-the-shelf'' SciBERT does not take global inter-document information into account. This is because SciBERT, like other pretrained language models, is trained via language modeling objectives, which only predict words or sentences given their in-document, nearby textual context. In contrast, we propose to incorporate citations into the model as a signal of inter-document relatedness, while still leveraging the model's existing strength in modeling language. 

\subsection{Citation-Based Pretraining Objective}

A citation from one document to another suggests that the documents are related.  To encode this relatedness signal into our representations, we design a loss function that trains the Transformer model to learn closer representations for papers when one cites the other, and more distant representations otherwise.
The high-level overview of the model is shown in Figure \ref{fig:overview}.  

In particular, each training instance is a triplet of papers: a query paper $\query$, a positive paper $\pos$ and a negative paper $\negative$. The positive paper is a paper that the query paper cites, and the negative paper is a paper that is not cited by the query paper (but that may be cited by $\pos$). We then train the model using the following triplet margin loss function:
\begin{equation}
\footnotesize
    \calL=
    \max\Big\{\Big(\dis\big(\query, \pos\big) - \dis\big(\query,\negative\big)+ m\Big),0\Big\} 
\label{eq:loss}
\end{equation}
\noindent where $d$ is a distance function and $m$ is the loss margin hyperparameter (we empirically choose $m=1$).
Here, we use the L2 norm distance:
$$ \dis(\calP^{A}, \calP^{B})=\lVert \V{v}_{A} - \V{v}_{B}\lVert_2 ,$$

\noindent where $\V{v}_{A}$ is the vector corresponding to the pooled output of the Transformer run on paper $A$ (Equation \ref{eq:forward-func}).\footnote{We also experimented with other distance functions (e..g, normalized cosine), but they underperformed the L2 loss.}
Starting from the trained SciBERT model, we pretrain the Transformer parameters on the citation objective to learn paper representations that capture document relatedness. 

\subsection{Selecting Negative Distractors}
\label{subsec:negative-sampling}

\setlength{\abovedisplayskip}{\parskip}
\setlength{\belowdisplayskip}{\parskip}

The choice of negative example papers $\negative $ is important when training the model. We consider two sets of negative examples: the first set simply consists of randomly selected papers from the corpus. Given a query paper, intuitively we would expect the model to be able to distinguish between cited papers, and uncited papers sampled randomly from the entire corpus. This inductive bias has been also found to be effective in content-based citation recommendation applications \cite{citeomatic}. But, random negatives may be easy for the model to distinguish from the positives.  To provide a more nuanced training signal, we augment the randomly drawn negatives with a more challenging second set of negative examples. We denote as ``hard negatives'' the papers that are not cited by the query paper, but {\em are} cited by a paper cited by the query paper, i.e. if $\calP^1\xrightarrow{cite}\calP^2$ and $\calP^2\xrightarrow{cite}\calP^3$ but $\calP^1\centernot{\xrightarrow{cite}}\calP^3$, then $\calP^3$ is a candidate hard negative example for $\calP^1$. We expect the hard negatives to be somewhat related to the query paper, but typically less related than the cited papers.  As we show in our experiments (\S\ref{sec:analysis}), including hard negatives results in more accurate embeddings compared to using random negatives alone.

\subsection{Inference}
At inference time, the model receives one paper, $\calP$, and it outputs the \sys's Transfomer pooled output activation as the paper representation for $\calP$ (Equation \ref{eq:forward-func}).  We note that for inference, \sys\ requires only the title and abstract of the given input paper; the model does not need any citation information about the input paper.  This means that \sys\ can produce embeddings even for new papers that have yet to be cited, which is critical for applications that target recent scientific papers.

%% file: 3-exp.tex
\section{\dataset\ Evaluation Framework}
\label{sec:eval}

Previous evaluations of scientific document representations in the literature tend to focus on small datasets over a limited set of tasks, and extremely high (99\%+) AUC scores are already possible on these data for English documents \cite{Chen2019ImprovingTN, Wang2019improving}. New, larger and more diverse benchmark datasets are necessary.  Here, we introduce a new comprehensive evaluation framework to measure the effectiveness of scientific paper embeddings, which we call \dataset. The framework consists of diverse tasks, ranging from citation prediction, to prediction of user activity, to document classification and paper recommendation. 
Note that \sys will not be further fine-tuned on any of the tasks; we simply plug in the embeddings as features for each task.
Below, we describe each of the tasks in detail and the evaluation data associated with it. In addition to our training data, we release all the datasets associated with the evaluation tasks. 

\subsection{Document Classification}
An important test of a document-level embedding is whether it is predictive of the class of the document.  Here, we consider two classification tasks in the scientific domain:

\paragraph{MeSH Classification \hspace{1em}}
In this task, the goals is to classify scientific papers according to their Medical Subject Headings (MeSH) \cite{lipscomb2000medical}.\footnote{\url{https://www.nlm.nih.gov/mesh/meshhome.html}} We construct a dataset consisting of 23K academic medical papers, where each paper is assigned one of 11 top-level disease classes such as cardiovascular diseases, diabetes, digestive diseases derived from the MeSH vocabulary.
The most populated category is Neoplasms (cancer) with 5.4K instances (23.3\% of the total dataset) while the category with least number of samples is Hepatitis (1.7\% of the total dataset). We follow the approach of \citet{feldman2019medical} in mapping the MeSH vocabulary to the disease classes. 

\paragraph{Paper Topic Classification  \hspace{1em}}
This task is predicting the topic associated with a paper using the predefined topic categories of the Microsoft Academic Graph (MAG)~\cite{Sinha2015AnOO}\footnote{\url{https://academic.microsoft.com/}}. MAG provides a database of papers, each tagged with a list of topics. The topics are organized in a hierarchy of 5 levels, where level 1 is the most general and level 5 is the most specific. For our evaluation, we derive a document classification dataset from the level 1 topics, where a paper is labeled by its corresponding level 1 MAG topic.
We construct a dataset of 25K papers, almost evenly split over the 19 different classes of level 1 categories in MAG.

\subsection{Citation Prediction}

As argued above, citations are a key signal of relatedness between papers.  We test how well different paper representations can reproduce this signal through citation prediction tasks. In particular, we focus on two sub-tasks: predicting \textit{direct citations}, and predicting \textit{co-citations}.
We frame these as ranking tasks and evaluate performance using \map and \ndcg, standard ranking metrics. 

\paragraph{Direct Citations} 
In this task, the model is asked to predict which papers are cited by a given query paper from a given set of candidate papers. The evaluation dataset includes approximately 30K total papers from a held-out pool of papers, consisting of 1K query papers and a candidate set of up to 5 cited papers and 25 (randomly selected) uncited papers. The task is to rank the cited papers higher than the uncited papers. For each embedding method, we require only comparing the L2 distance between the raw embeddings of the query and the candidates, without any additional trainable parameters. 

\paragraph{Co-Citations}
This task is similar to the \textit{direct citations} but instead of predicting a cited paper, the goal is to predict a highly co-cited paper with a given paper. Intuitively, if papers A and B are cited frequently together by several papers, this shows that the papers are likely highly related and a good paper representation model should be able to identify these papers from a given candidate set. The dataset consists of 30K total papers and is constructed similar to the \textit{direct citations} task.

\subsection{User Activity}
The embeddings for similar papers should be close to each other; 
we use user activity as a proxy for identifying similar papers and test the model's ability to recover this information. 
Multiple users consuming the same items as one another is a classic relatedness signal and forms the foundation for recommender systems and other applications \cite{schafer2007collaborative}. In our case, we would expect that when users look for academic papers, the papers they view in a single browsing session tend to be related.  Thus, accurate paper embeddings should, all else being equal, be relatively more similar for papers that are frequently viewed in the same session than for other papers.  
To build benchmark datasets to test embeddings on user activity, we obtained logs of user sessions from a major academic search engine. We define the following two tasks on which we build benchmark datasets to test embeddings:

\paragraph{Co-Views}
Our co-views dataset consists of approximately 30K papers.  To construct it, we take 1K random papers that are not in our train or development set and associate with each one up to 5 frequently co-viewed papers and 25 randomly selected papers (similar to the approach for citations). Then, we require the embedding model to rank the co-viewed papers higher than the random papers by comparing the L2 distances of raw embeddings.  We evaluate performance using standard ranking metrics, \ndcg and \map.

\paragraph{Co-Reads}
If the user clicks to access the PDF of a paper from the paper description page, this is a potentially stronger sign of interest in the paper.  In such a case we assume the user will read at least parts of the paper and refer to this as a ``read'' action.  Accordingly, we define a ``co-reads'' task and dataset analogous to the co-views dataset described above. This dataset is also approximately 30K papers.

\subsection{Recommendation}

In the recommendation task, we evaluate the ability of paper embeddings to boost performance in a production recommendation system.  Our recommendation task aims to help users navigate the scientific literature by ranking a set of ``similar papers'' for a given paper.  
We use a dataset of user clickthrough data for this task which consists of 22K clickthrough events from a public scholarly search engine. %
We partitioned the examples temporally into train (20K examples), validation (1K), and test (1K) sets.  As is typical in clickthrough data on ranked lists, the clicks are biased toward the top of original ranking presented to the user.  To counteract this effect, we computed propensity scores using a swap experiment \cite{Agarwal2019EstimatingPB}.  The propensity scores give, for each position in the ranked list, the relative frequency that the position is over-represented in the data due to exposure bias.  We can then compute de-biased evaluation metrics by dividing the score for each test example by the propensity score for the clicked position.  We report propensity-adjusted  versions of the standard ranking metrics Precision@1 ($\hat{\mathrm{P}@1}$) and Normalized Discounted Cumulative Gain ($\hat{\mathrm{n}\textsc{dcg}}$).

We test different embeddings on the recommendation task by including cosine embedding distance\footnote{Embeddings are L2 normalized and in this case cosine distance is equivalent to L2 distance.} as a feature within an existing recommendation system that includes several other informative features (title/author similarity, reference and citation overlap, etc.). Thus, the recommendation experiments measure whether the embeddings can boost the performance of a strong baseline system on an end task.  For \sys , we also perform an online A/B test to measure whether its advantages on the offline dataset translate into improvements on the online recommendation task (\S\ref{sec:results}).

\section{Experiments}
\label{sec:exp-details}

\setlength{\dashlinedash}{0.2pt}
\setlength{\dashlinegap}{1pt}
\setlength{\arrayrulewidth}{0.4pt}

\paragraph{Training Data}

To train our model, we use a subset of the Semantic Scholar corpus \cite{Ammar2018ConstructionOT} consisting of about 146K query papers (around 26.7M tokens) with their corresponding outgoing citations, and we use an additional 32K papers for validation.
For each query paper we construct up to 5 training triples comprised of a query, a positive, and a negative paper. The positive papers are sampled from the direct citations of the query, while negative papers are chosen either randomly or from citations of citations (as discussed in \S\ref{subsec:negative-sampling}). We empirically found it helpful to use 2 hard negatives (citations of citations) and 3 easy negatives (randomly selected papers) for each query paper. This process results in about 684K training triples and 145K validation triples.

\paragraph{Training and Implementation}
We implement our model in AllenNLP \cite{gardner-etal-2018-allennlp}. We initialize the model from SciBERT pretrained weights \cite{Beltagy2019SciBERT} since it is the state-of-the-art pretrained language model on scientific text. We continue training all model parameters on our training objective (Equation \ref{eq:loss}). We perform minimal tuning of our model's hyperparameters based on the performance on the validation set, while baselines are extensively tuned. Based on initial experiments, we use a margin $m{=}1$ for the triplet loss. For training, we use the Adam optimizer \cite{Kingma2014AdamAM} following the suggested hyperparameters in \citet{Devlin2018BERTPO} (LR: 2e-5, Slanted Triangular LR scheduler\footnote{Learning rate linear warmup followed by linear decay.}~\cite{howard-ruder-2018-universal} with number of train steps equal to training instances and cut fraction of 0.1). We train the model on a single Titan V GPU (12G memory) for 2 epochs, with batch size of 4 (the maximum that fit in our GPU memory) and use gradient accumulation for an effective batch size of 32. Each training epoch takes approximately 1-2 days to complete on the full dataset. We release our code and data to facilitate reproducibility.
\footnote{\url{https://github.com/allenai/specter}}

\paragraph{Task-Specific Model Details}
For the classification tasks, we used a linear SVM  where embedding vectors were the only features. The $C$ hyper-parameter was tuned via a held-out validation set.  For the recommendation tasks, we use a feed-forward ranking neural network that takes as input ten features designed to capture the similarity between each query and candidate paper, including the cosine similarity between the query and candidate embeddings and manually-designed features computed from the papers' citations, titles, authors, and publication dates.

\begin{table*}[t]
\centering
\footnotesize
\setlength{\tabcolsep}{4.9pt}
\renewcommand{\arraystretch}{1.2}
\begin{tabular}{@{}lrrrrrrrrrrrrr@{}}
\toprule
Task $\rightarrow$                 & \multicolumn{2}{c}{Classification} & \multicolumn{4}{c}{User activity prediction}                        & \multicolumn{4}{c}{Citation prediction}                & \multicolumn{2}{c}{\multirow{2}{*}{\begin{tabular}{@{}c@{}}Recomm.\end{tabular}}} & 
\multirow{3}{*}{\begin{tabular}{@{}c@{}}Avg.\end{tabular}}
\\ \cmidrule(lr){2-3} \cmidrule(lr){4-7} \cmidrule(lr){8-11} 
Subtask $\rightarrow$              & MAG              & MeSH            & \multicolumn{2}{c}{Co-View} & \multicolumn{2}{c}{Co-Read} & \multicolumn{2}{c}{Cite} & \multicolumn{2}{c}{Co-Cite} & \multicolumn{2}{c}{} &                   \\ \cdashline{1-13}
Model $\downarrow$ / Metric $\rightarrow$               & F1               & F1              & \map        & \ndcg        & \map         & \ndcg        & \map       & \ndcg       & \map         & \ndcg        & $\hat{\mathrm{n}\textsc{dcg}}$        & $\hat{\mathrm{P}@1}$       &  \\ \midrule
Random                             & 4.8  & 9.4  & 25.2 & 51.6 & 25.6 & 51.9 & 25.1 & 51.5 & 24.9 & 51.4 & 51.3 & 16.8 & 32.5 \\ 
Doc2vec \citeyearpar{doc2vec}      & 66.2 & 69.2 & 67.8 & 82.9 & 64.9 & 81.6 & 65.3 & 82.2 & 67.1 & 83.4 & 51.7 & 16.9 & 66.6 \\
Fasttext-sum \citeyearpar{fasttext}        & 78.1 & 84.1 & 76.5 & 87.9 & 75.3 & 87.4 & 74.6 & 88.1 & 77.8 & 89.6 & 52.5 & 18.0 & 74.1 \\
SIF \citeyearpar{sif}                  & 78.4 & 81.4 & 79.4 & 89.4 & 78.2 & 88.9 & 79.4 & 90.5 & 80.8 & 90.9 & 53.4 & 19.5 & 75.9 \\
ELMo \citeyearpar{elmo}                 & 77.0 & 75.7 & 70.3 & 84.3 & 67.4 & 82.6 & 65.8 & 82.6 & 68.5 & 83.8 & 52.5 & 18.2 & 69.0 \\
Citeomatic \citeyearpar{citeomatic}           & 67.1 & 75.7 & 81.1 & 90.2 & 80.5 & 90.2 & 86.3 & 94.1 & 84.4 & 92.8 & 52.5 & 17.3 & 76.0 \\ 
SGC \citeyearpar{sgc}               & 76.8 & 82.7 & 77.2 & 88.0 & 75.7 & 87.5 & \bf{91.6} & \bf{96.2} & 84.1 & 92.5 & 52.7 & 18.2 & 76.9 \\
SciBERT \citeyearpar{Beltagy2019SciBERT} & 79.7 & 80.7 & 50.7 & 73.1 & 47.7 & 71.1 & 48.3 & 71.7 & 49.7 & 72.6 & 52.1 & 17.9 & 59.6 \\ 
Sent-BERT \citeyearpar{sentence_bert}           & 80.5 & 69.1 & 68.2 & 83.3 & 64.8 & 81.3 & 63.5 & 81.6 & 66.4 & 82.8 & 51.6 & 17.1 & 67.5 \\ \hdashline
\sys (Ours)          &  \bf{82.0} &	\bf{86.4} & \bf{83.6} & \bf{91.5} & \bf{84.5} & \bf{92.4} & 88.3 &	94.9 & \bf{88.1} & \bf{94.8} & \bf{53.9}& \bf{20.0} & \bf{80.0} \\
\bottomrule
\end{tabular}
\caption{Results on the \dataset evaluation suite consisting of 7 tasks. 
}
\label{tab:results}
\end{table*}

\paragraph{Baseline Methods}

Our work falls into the intersection of textual representation, citation mining, and graph learning, and we evaluate against state-of-the-art baselines from each of these areas. 
We compare with several strong textual models: SIF \cite{sif}, a method for learning document representations by removing the first principal component of aggregated word-level embeddings which we pretrain on scientific text; SciBERT \cite{Beltagy2019SciBERT} a state-of-the-art pretrained Transformer LM for scientific text; and Sent-BERT \cite{sentence_bert}, a model that uses negative sampling to tune BERT for producing optimal sentence embeddings. We also compare with Citeomatic \cite{citeomatic}, a closely related paper representation model for citation prediction which trains content-based representations with citation graph information via dynamically sampled triplets, and SGC \cite{sgc}, a state-of-the-art graph-convolutional approach.
For completeness, additional baselines are also included; due to space constraints we refer to Appendix \ref{appendix:a} for detailed discussion of all baselines. We tune hyperparameters of baselines to maximize performance on a separate validation set.

%% file: 4-results.tex
\section{Results}
\label{sec:results}

Table~\ref{tab:results} presents the main results corresponding to our evaluation tasks (described in \S\ref{sec:eval}).
Overall, we observe substantial improvements across all tasks with average performance of 80.0 across all metrics on all tasks which is a 3.1 point absolute improvement over the next-best baseline. We now discuss the results in detail.

For document classification, we report macro F1, a standard classification metric. We observe that the classifier performance when trained on our representations is better than when trained on any other baseline. Particularly, on the MeSH (MAG) dataset, we obtain an 86.4 (82.0) F1 score which is about a $\Delta{=} +2.3$ ($+1.5$) point absolute increase over the best baseline on each dataset respectively. 
Our evaluation of the learned representations on predicting user activity is shown in the ``User activity'' columns of Table \ref{tab:results}. 
\sys achieves a \map score of 83.8 on the co-view task, and 84.5 on co-read, improving over the best baseline (Citeomatic in this case) by 2.7 and 4.0 points, respectively.
We observe similar trends for the ``citation'' and ``co-citation'' tasks, with our model outperforming virtually all other baselines except for SGC, which has access to the citation graph at training and test time.\footnote{For SGC, we remove development and test set citations and co-citations during training.  We also remove incoming citations from development and test set queries as these would not be available at test time in production.} Note that methods like SGC cannot be used in real-world setting to embed new papers that are not cited yet. 
On the other hand, on co-citation data our method is able to achieve the best results with \ndcg of 94.8, improving over SGC with 2.3 points. Citeomatic also performs well on the citation tasks, as expected given that its primary design goal was citation prediction. Nevertheless, our method slightly outperforms Citeomatic on the direct citation task, while substantially outperforming it on co-citations (+2.0 \ndcg).

Finally, for recommendation task, we observe that \sys outperforms all other models on this task as well, with \ndcg of 53.9. On the recommendations task, as opposed to previous experiments, the differences in method scores are generally smaller. This is because for this task the embeddings are used along with several other informative features in the ranking model (described under task-specific models in \S\ref{sec:exp-details}), meaning that embedding variants have less opportunity for impact on overall performance.

We also performed an online study to evaluate whether \sys\ embeddings offer similar advantages in a live application.  We performed an online A/B test comparing our \sys -based recommender to an existing production recommender system for similar papers that ranks papers by a textual similarity measure.  In a dataset of 4,113 clicks, we found that \sys ranker improved clickthrough rate over the baseline by 46.5\%, demonstrating its superiority.

We emphasize that our citation-based pretraining objective is critical for the performance of \sys ; removing this and using a vanilla SciBERT results in decreased performance on all tasks.

%% file: 5-analysis.tex
\section{Analysis}
\label{sec:analysis}

In this section, we analyze several design decisions in \sys , provide a visualization of its embedding space, and experimentally compare \sys 's use of fixed embeddings against a fine-tuning approach.

\paragraph{Ablation Study}

We start by analyzing how adding or removing metadata fields from the input to \sys alters performance.  The results are shown in the top four rows of Table \ref{tab:ablation} (for brevity, here we only report the average of the metrics from each task). 
We observe that removing the abstract from the textual input and relying only on the title results in a substantial decrease in performance.
More surprisingly, adding authors as an input (along with title and abstract) hurts performance.\footnote{We experimented with both concatenating authors with the title and abstract and also considering them as an additional field. Neither were helpful.} One possible explanation is that author names are sparse in the corpus, making it difficult for the model to infer document-level relatedness from them. 
As another possible reason of this behavior, tokenization using Wordpieces might be suboptimal for author names. Many author names are  out-of-vocabulary for SciBERT and thus, they might be split into sub-words and shared across names that are not semantically related, leading to noisy correlation.
Finally, we find that adding venues slightly decreases performance,\footnote{Venue information in our data came directly from publisher provided metadata and thus was not normalized. Venue normalization could help improve results.} except on document classification (which makes sense, as we would expect venues to have high correlation with paper topics). The fact that \sys does not require inputs like authors or venues makes it applicable in situations where this metadata is not available, such as matching reviewers with anonymized submissions, or performing recommendations of anonymized preprints (e.g., on OpenReview).

\begin{table}[t]
\footnotesize
\centering
\setlength{\tabcolsep}{5pt}
\renewcommand{\arraystretch}{1.2}
\begin{tabular}{@{}lrrrrr@{}}
\toprule
 &CLS       & USR   & CITE  & REC  & Avg.      \\ \midrule
\sys      & 84.2 & \bf{88.4} & \bf{91.5}  & \bf{36.9} & \bf{80.0} \\ \hdashline
$-$ abstract & 82.2 & 72.2 & 73.6  & 34.5 & 68.1 \\
$+$ venue    & \bf{84.5} & 88.0 & 91.2  & 36.7 & 79.9\\
$+$ author   & 82.7 & 72.3 & 71.0  & 34.6 & 67.3\\ \hdashline
No hard negatives & 82.4	& 85.8 &	89.8 &	36.8 & 78.4\\ 
Start w/ BERT-Large   & 81.7 &	85.9 &	87.8 &	36.1 & 77.5 \\
\bottomrule
\end{tabular}
\caption{Ablations: Numbers are averages of metrics for each evaluation task: CLS: classification, USR: User activity, CITE: Citation prediction, REC: Recommendation, Avg. average over all tasks \& metrics.}
\label{tab:ablation}
\end{table}

One design decision in \sys\ is to use a set of hard negative distractors in the citation-based fine-tuning objective.  The fifth row of Table \ref{tab:ablation} shows that this is important---using only easy negatives reduces performance on all tasks. 
While there could be other potential ways to include hard negatives in the model, our simple approach of including citations of citations is effective. 
The sixth row of the table shows that using a strong {\em general-domain} language model (BERT-Large) instead of SciBERT in \sys\ reduces performance considerably. This is reasonable because unlike BERT-Large, SciBERT is pretrained on scientific text.

\paragraph{Visualization}

\begin{figure}
\centering
\begin{subfigure}{.5\linewidth}
  \centering
  \includegraphics[width=\linewidth]{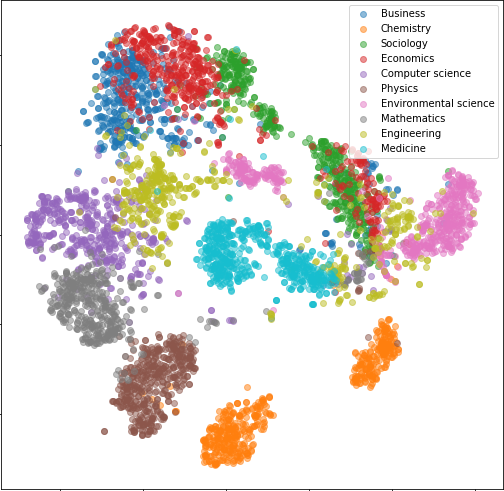}
  \caption{\sys}
  \label{fig:sub1}
\end{subfigure}%
\begin{subfigure}{.5\linewidth}
  \centering
  \includegraphics[width=\linewidth]{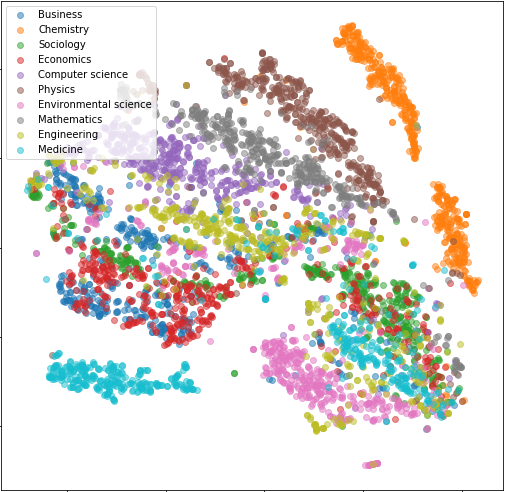}
  \caption{SciBERT}
  \label{fig:sub2}
\end{subfigure}
\caption{t-SNE visualization of paper embeddings and their corresponding MAG topics.}
\label{fig:tsne}
\vspace{-8pt}
\end{figure}

Figure~\ref{fig:tsne} shows t-SNE~\cite{Maaten2014AcceleratingTU} projections of our embeddings (\sys) compared with the SciBERT baseline for a random set of papers. 
When comparing \sys embeddings with SciBERT, we observe that our embeddings are better at encoding topical information, as the clusters seem to be more compact. Further, we see some examples of cross-topic relatedness reflected in the embedding space (e.g., 
Engineering, Mathematics and Computer Science are close to each other, while Business and Economics are also close to each other). 
To quantify the comparison of visualized embeddings in Figure \ref{fig:tsne}, we use the DBScan clustering algorithm \cite{ester1996density} on this 2D projection. We use the completeness and homogeneity clustering quality measures introduced by \citet{rosenberg2007v}.
For the points corresponding to Figure~\ref{fig:tsne}, the homogeneity and completeness values for \sys are respectively 0.41 and 0.72 compared with SciBERT's 0.19 and 0.63, a clear improvement on separating topics using the projected embeddings.

\paragraph{Comparison with Task Specific Fine-Tuning}
While the fact that \sys does not require fine-tuning makes its paper embeddings less costly to use, often the best performance from pretrained Transformers is obtained when the models are fine-tuned directly on each end task.  We experiment with fine-tuning SciBERT on our tasks, and find this to be generally inferior to using our fixed representations from \sys.  Specifically, we fine-tune SciBERT directly on task-specific signals instead of citations. To fine-tune on task-specific data (e.g., user activity), we used a dataset of co-views with 65K query papers, co-reads with 14K query papers, and co-citations (instead of direct citations) with 83K query papers. As the end tasks are ranking tasks, for all datasets we construct up to 5 triplets and fine-tune the model using triplet ranking loss. The positive papers are sampled from the most co-viewed (co-read, or co-cited) papers corresponding to the query paper. We also include both easy and hard distractors as when training \sys (for hard negatives we choose the least non-zero co-viewed (co-read, or co-cited) papers). We also consider training jointly on all task-specific training data sources in a multitask training process, where the model samples training triplets from a distribution over the sources. 
As illustrated in Table \ref{tab:coview}, without any additional final task-specific fine-tuning, \sys still outperforms a SciBERT model fine-tuned on the end tasks as well as their multitask combination, further demonstrating the effectiveness and versatility of \sys embeddings.\footnote{We also experimented with further task-specific fine-tuning of our \sys on the end tasks but we did not observe additional improvements.}

\begin{table}[t]
\footnotesize
\centering
\setlength{\tabcolsep}{1.2pt}
\begin{tabular}{@{}lrrrrr@{}}
\toprule
Training signal & CLS       & USR   & CITE  & REC  & All     \\ \midrule
\sys      & 84.2 & \bf{88.4} & \bf{91.5}  & \bf{36.9} & \bf{80.0} \\ \hdashline
SciBERT fine-tune on co-view   & 83.0 &	84.2 &	84.1 &	36.4 & 76.0\\
SciBERT fine-tune on co-read    & 82.3 &	85.4 &	86.7 &	36.3 & 77.1\\ 
SciBERT fine-tune on co-citation & 
82.9 &	84.3 &	85.2	& 36.6 & 76.4\\
SciBERT fine-tune on multitask & 83.3 &	86.1 &	88.2 &	36.0	& 78.0\\ \bottomrule
\end{tabular}
\caption{Comparison with task-specific fine-tuning.}
\label{tab:coview}
\vspace{-8pt}
\end{table}

%% file: 6-related.tex
\section{Related Work}
\label{sec:related}

Recent representation learning methods in NLP rely on training large neural language models on unsupervised data \cite{elmo,radford2018improving,Devlin2018BERTPO,Beltagy2019SciBERT,Liu2019RoBERTaAR}. 
While successful at many sentence- and token-level tasks, our focus is on using the models for document-level representation learning, which has remained relatively under-explored.

There have been other efforts in document representation learning such as extensions of word vectors to documents \cite{doc2vec,doc2sent2vec,extended_paragraph_vector,wme,nvsm}, convolution-based methods \cite{cnn_embed,snrm}, and variational autoencoders \cite{savae,Wang2019improving}. 
Relevant to document embedding, sentence embedding is a relatively well-studied area of research. Successful approaches include seq2seq models \cite{Kiros2015SkipThoughtV}, BiLSTM Siamese networks \cite{williams-etal-2018-lstm-sent}, leveraging supervised data 
from other corpora \cite{conneau-etal-2017-supervised}, and using discourse relations \cite{nie-etal-2019-dissent}, and BERT-based methods \cite{sentence_bert}. Unlike our proposed method, the majority of these approaches do not consider any notion of inter-document relatedness when embedding documents.

Other relevant work combines textual features with network structure \cite{Tu2017CANECN, Zhang2018DiffusionMF,citeomatic,Shen2018ImprovedSN,Chen2019ImprovingTN,Wang2019improving}. These works typically do not leverage the recent pretrained contextual representations and with a few exceptions such as the recent work by \citet{Wang2019improving}, they cannot generalize to unseen documents like our \sys approach. Context-based citation recommendation is another related application where models rely on citation contexts \cite{Jeong2019ACC} to make predictions. These works are orthogonal to ours as the input to our model is just paper title and abstract. 
Another related line of work is graph-based representation learning methods \cite{bruna2013spectral,kipf2016semi,hamilton2017inductive,graphsage,sgc,wu2019comprehensive}. 
Here, we compare to a graph representation learning model, SGC (Simple  Graph  Convolution) \cite{sgc}, which is a state-of-the-art graph convolution approach for representation learning. 
\sys uses pretrained language models in combination with graph-based citation signals, which enables it to outperform the graph-based approaches in our experiments. %

\sys embeddings are based on only the title and abstract of the paper.  Adding the full text of the paper would provide a more complete picture of the paper's content and could improve accuracy \cite{Cohen2010TheSA,Lin2008IsSF,Schuemie2004DistributionOI}.  However,
the full text of many academic papers is not freely available.  Further, modern language models have strict memory limits on input size, which means new techniques would be required in order to leverage the entirety of the paper within the models.  Exploring how to use the full paper text within \sys\ is an item of future work.

Finally, one pain point in academic paper recommendation research has been a lack of publicly available datasets \cite{chen2018research,Kanakia2019ASH}. To address this challenge, we release \dataset, our evaluation benchmark which includes an anonymized clickthrough dataset from an online recommendations system.

%% file: 9-appendix.tex
\section{Appendix A - Baseline Details}
\label{appendix:a}

\setlist[enumerate]{wide=0pt}
\begin{enumerate}
\setlength\itemsep{0pt}

\item \textbf{Random}
Zero-mean 25-dimensional vectors were used as representations for each document.

\item \textbf{Doc2Vec}
Doc2Vec is one of the earlier neural document/paragraph representation methods \cite{doc2vec}, and is a natural comparison. We trained Doc2Vec on our training subset using Gensim~\cite{gensim}, and chose the hyperparameter grid using suggestions from~\citet{doc2vec_hyperparam}. The hyperparameter grid used:
\begin{verbatim}
{'window': [5, 10, 15],
 'sample': [0, 10 ** -6, 10 ** -5],
 'epochs': [50, 100, 200]},
\end{verbatim}
for a total of 27 models.  The other parameters were set as follows: \verb|vector_size=300|, \verb|min_count=3|, \verb|alpha=0.025|, \verb|min_alpha=0.0001|, \verb|negative=5|, \verb|dm=0|, \verb|dbow=1|, \verb|dbow_words=0|.

\item \textbf{Fasttext-Sum}
This simple baseline is a weighted sum of pretrained word vectors. We trained our own 300 dimensional fasttext embeddings \cite{fasttext} on a corpus of around 3.1B tokens from scientific papers which is similar in size to the SciBERT corpus \cite{Beltagy2019SciBERT}. We found that these pretrained embeddings substantially outperform alternative off-the-shelf embeddings. We also use these embeddings in other baselines that require pretrained word vectors (i.e., SIF and SGC that are described below). The summed bag of words representation has a number of weighting options, which are extensively tuned on a validation set for best performance.

\item \textbf{SIF}
The SIF method of \citet{sif} is a strong text representation baseline that takes a weighted sum of pretrained word vectors (we use fasttext embeddings described above), then computes the first principal component of the document embedding matrix and subtracts out each document embedding's projection to the first principal component.

We used a held-out validation set to choose $a$ from the range [1.0e-5, 1.0e-3] spaced evenly on a log scale. The word probability $p(w)$ was estimated on the training set only. When computing term-frequency values for SIF, we used scikit-learn's TfidfVectorizer with the same parameters as enumerated in the preceding section. \verb|sublinear_tf|, \verb|binary|, \verb|use_idf|, \verb|smooth_idf| were all set to \verb|False|. Since SIF is a sum of pretrained fasttext vectors, the resulting dimensionality is 300.

\item \textbf{ELMo}
ELMo \cite{elmo} provides contextualized representations of tokens in a document. It can provide paragraph or document embeddings by averaging each token's representation for all 3 LSTM layers. We used the 768-dimensional pretrained ELMo model in AllenNLP \cite{gardner-etal-2018-allennlp}.

\item \textbf{Citeomatic}
The most relevant baseline is Citeomatic \cite{citeomatic}, which is an academic paper representation model that is trained on the citation graph via sampled triplets. Citeomatic representations are an L2 normalized weighted sum of title and abstract embeddings, which are trained on the citation graph with dynamic negative sampling. Citeomatic embeddings are 75-dimensional. 

\item \textbf{SGC}
Since our algorithm is trained on data from the citation graph, we also compare to a state-of-the-art graph representation learning model: SGC (Simple Graph Convolution) \cite{sgc}, which is a graph convolution network. 
An alternative comparison would have been GraphSAGE \cite{graphsage}, but SGC (with no learning) outperformed an unsupervised variant of GraphSAGE on the Reddit dataset\footnote{There were no other direct comparisons in \citet{sgc}},
Note that SGC with no learning boils down to graph propagation on node features (in our case nodes are academic documents). Following \citet{hamilton2017inductive}, we used SIF features as node representations, and applied SGC with a range of parameter $k$, which is the number of times the normalized adjacency is multiplied by the SIF feature matrix. Our range of $k$ was 1 through 8 (inclusive), and was chosen with a validation set. For the node features, we chose the SIF model with $a=0.0001$, as this model was observed to be a high-performing one. This baseline is also 300 dimensional.
 
 \item \textbf{SciBERT}
To isolate the advantage of \sys 's citation-based fine-tuning objective, we add a controlled comparison with SciBERT \cite{Beltagy2019SciBERT}. Following \citet{Devlin2018BERTPO} we take the last layer hidden state corresponding to the \cls token as the aggregate document representation.\footnote{We also tried the alternative of averaging all token representations, but this resulted in a slight performance decrease compared with the \cls pooled token.}

\item \textbf{Sentence BERT}
Sentence BERT \cite{sentence_bert} is a general-domain pretrained model aimed at embedding sentences. The authors fine-tuned BERT using a triplet loss, where positive sentences were from the same document section as the seed sentence, and distractor sentences came from other document sections. The model is designed to encode sentences as opposed to paragraphs, so we embed the title and each sentence in the abstract separately, sum the embeddings, and L2 normalize the result to produce a final 768-dimensional paper embedding.\footnote{We used the `bert-base-wikipedia-sections-mean-tokens' model released by the authors: \url{https://github.com/UKPLab/sentence-transformers}}

During hyperparameter optimization we chose how to compute TF and IDF values weights by taking the following non-redundant combinations of scikit-learn's TfidfVectorizer \cite{scikit-learn} parameters: \verb|sublinear_tf|, \verb|binary|, \verb|use_idf|, \verb|smooth_idf|. There were a total of 9 parameter combinations. The IDF values were estimated on the training set.  The other parameters were set as follows: \verb|min_df=3|, \verb|max_df=0.75|, \verb|strip_accents='ascii'|, \verb|stop_words='english'|,  \verb|norm=None|, \verb|lowercase=True|. For training of fasttext, we used all default parameters with the exception of setting dimension to 300 and \verb|minCount| was set to 25 due to the large corpus.

\end{enumerate}

\label{appendix:b}